# Spike-and-wave epileptiform discharge pattern detection based on Kendall's Tau-b coefficient

Antonio QUINTERO-RINCÓN[1,*], Catalina CARENZO[1,+], Joaquín EMS[1,+], Lourdes HIRSCHSON[1,+], Valeria MURO[2] and Carlos D´GIANO[2]

[1] Department of Bioengineering, Instituto Tecnológico de Buenos Aires, Avenida Eduardo Madero 399, C1106ACD CABA, Buenos Aires, Argentina.
[2] Centro Integral de Epilepsia, Fundación para la Lucha contra las Enfermedades Neurológicas de la Infancia (FLENI), Montañeses 2325, C1428 AQK, Buenos Aires.
E-mail(s): ccarenzo@itba.edu.ar; jems@itba.edu.ar; lhirschson@itba.edu.ar; vmuro@fleni.org.ar; cdgiano@fleni.org.ar

* Author to whom correspondence should be addressed; Tel.:+54011 2150-4800.

[+] All authors contributed equally to this work.

**Abstract**

Epilepsy is an important public health issue. An appropriate epileptiform discharge pattern detection of this neurological disease is a typical problem in biomedical engineering. In this paper, a new method is proposed for spike-and-wave discharge pattern detection based on Kendall's Tau-b coefficient. The proposed approach is demonstrated on a real dataset containing spike-and-wave discharge signals, where our performance is evaluated in terms of high Specificity, rule in (SpPIn) with 94% for patient-specific spike-and-wave discharge detection and 83% for a general spike-and-wave discharge detection.

## Introduction

Electroencephalography (EEG) is widely used to record the electrical activity of the brain in neurological health centers. EEGs help study and diagnose several different types of brain disorders, such as epilepsy. Epilepsy is a neurological disorder caused by intense activity of nerve cells in the brain, causing seizures. This uncontrolled electrical disturbance can cause changes in the levels of consciousness, behavior and body movements. A spike-and-wave discharge (SWD) is an epileptiform discharge with a regular and symmetric morphology, which typically starts and ends abruptly [1], see Figure 1. In current literature, epileptiform pattern recognition has extensive signal processing methods for accurate detection. Some approaches can be found using spectrogram with harmonic analysis [2], discrete cosine transform coupled with Daubechies wavelet [3], time-frequency and non-linear analysis [4], spatio-temporal analysis combine with autocovariance [5], Hilbert-Huang transform [6], Pseudo-Wigner-Ville and Choi-William distributions followed by Renyi's entropy [7], convolutional neural networks [8], statistical modeling [9, 10], bootstrap resampling [11], complex network of neuronal oscillators [12], or by using cross-approximate entropy [13]. See [14-16] for some studies in morphological similarity or concordance between signals and [17] for a complete state-of-the-art about methods of automated absence seizure detection.

Kendall's Tau-b coefficient is a nonparametric correlation analysis used to measure the ordinal association or concordance between two measured quantities. In our case this was done with two morphological waveforms signals. EEG studies with this coefficient are very diverse, in [18] with correlational neuronal activity in sleep; in [19] to compare the quality of life in epileptic patients; in [20] using time series or envelopes EEG´s pairs to quantify dependence between channels were investigated, or in [21] where connectivity assessed by intracranial electrical stimulation was used to quantify and detect the association among multichannel biosignals.

In this work we study Kendall's Tau-b coefficient between two morphological waveforms of the same length. The goal is to estimate the statistical relationship between a spike-and-wave epileptiform discharge (SWD) against each EEG segment by channel, in order to quantify and detect the morphological similarity or concordance between signals.

**Material and Method**

*Database*

A database with 780 monopolar 256 Hz signals was created measured from ten patients from Fundación Lucha contra las Enfermedades Neurológicas Infantiles (FLENI). 390 spike-and-wave signals had different time-length and waveform but their morphology is preserved, while 390 non-spike-and-wave signals had normal waveforms, see [9,10] for more details. In this work, only 300 spike-and-wave signals were used, see Figure 1 and Figure 3.

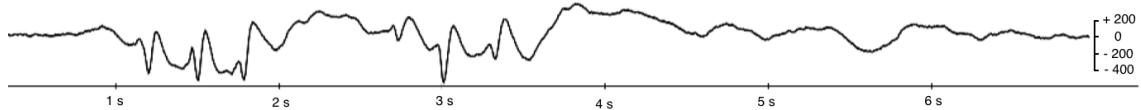

**Figure 1.** Spike-and-wave discharge example in an EEG channel. We can see the SWD epileptiform pattern between 1- 4 seconds

SWD are restricted to a narrow frequency band between 1-3 Hz. Each EEG was acquired with a 22-channel array using the standard 10-20 system through the following channels: Fp1, Fp2, F7, F3, Fz, F4, F8, T3, C3, Cz, C4, T4, T5, P3, Pz, P4, T6, O1, O2, Oz, FT10 and FT9, see Figure 2.

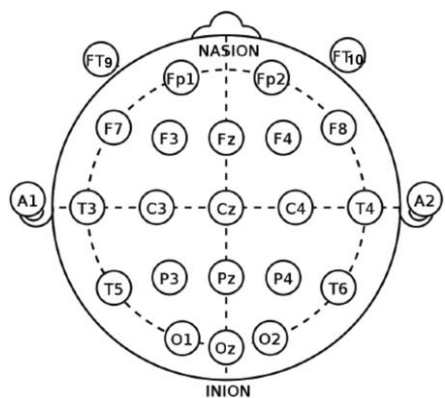

**Figure 2.** Electrodes position used in this work.

*Methodology*

Let $X_r \in \mathbb{R}^{M \times N}$ denote the matrix assembly M EEG signals $x_m \in \mathbb{R}^{1 \times N}$ measured simultaneously on different channels and at $N$ discrete time instants.

Let $Y_p \in \mathbb{R}^{1 \times L}$ denote the spike-and-wave vector with length $L$. The proposed methodology was composed of four stages. The first stage splits each channel of the original signal $X_r$ into a set of $N/L$ segments of the same sampling size of each spike-and-wave discharge without overlap. This segmentation was done using a rectangular sliding window, such that

$$\Omega_s = \frac{-(N/L) - 1}{2} \leq s \leq \frac{(N/L) - 1}{2} \tag{1}$$

so that $X_s = \Omega_s X_r$. The second stage consisted of filtering the segments $X_s$ and $Y_p$ with a $k$-point moving average filter, so that

$$\Gamma[n] = \sum_{k=0}^{K-1} \frac{1}{b[k]} \Gamma_i[n-k] \tag{2}$$

where $\Gamma[n] = X_f[n]$ for input $\Gamma_i = X_s$; $\Gamma[n] = Y_f[n]$ for input $\Gamma_i = Y_p$ and empirical value $k=5$, this value was chosen in order to not to change the original waveform signal characteristics. Next, in third stage $X_f[n]$ and $Y_f[n]$ were scaled between lower limit $\alpha = -1$ and upper limit $\beta = +1$ using

$$Y[n] = \frac{(v[n] - min(v[n]))(\beta - \alpha)}{max(v[n]) - min(v[n])} + \alpha \tag{3}$$

where $Y[n] = X[n]$ for input $v[n] = X_f[n]$ and $Y[n] = Y[n]$ for input $v[n] = Y_f[n]$.

Finally in the fourth stage a Kendall's Tau-b coefficient was applied between the SWD pattern $Y[n]$ and each $X[n]$ segment in order to detect a spike-and-wave epileptiform discharge pattern into EEG signals. We now introduce the Kendall's Tau-b coefficient used in this paper.

*Kendall's Tau-b Coefficient*

Let $X$ be each segment filtered and scaled for each EEG channel and $Y$ a SWD pattern filtered and scaled waveform, both through $\Gamma[n]$ and $Y[n]$, see equations (2) and (3). Then the bivariate couples $(x_1, y_1), \ldots, (x_n, y_n)$ are a sample of observations of the combined random variables $X$ and $Y$, such that all the values of $x_i$ and $y_i$ are unique. Two bivariate observations $(x_i, y_i)$ and $(x_j, y_j)$ are:
- Concordant: when $x_i > x_j$ and $y_i > y_j$, or $x_i < x_j$ and $y_i < y_j$
- Discordant: when $x_i > x_j$ and $y_i < y_j$, or $x_i < x_j$ and $y_i > y_j$
- Neither concordant: when $x - i = x - j$, or $y_i = y_j$

Kendall's Tau-b coefficient can be estimated in two ways, a) according the numbers of concordant $(n_c)$ and discordant $(n_d)$ pairs.

$$\tau = 2 \frac{n_c - n_d}{n(n-1)} \tag{4}$$

where the denominator is the total number of pair combinations, $-1 \leq \tau \leq 1$. Or b) through the following expression

$$\tau = \frac{2}{n(n-1)} \sum_{i<j}^{N/L} sgn(x_i - x_j) sgn(y_i - y_j) \tag{5}$$

where
- $\tau = +1$: Implies perfect agreement between two rankings, when $X$ and $Y$ are the same.
- $\tau = -1$: Implies perfect disagreement between two rankings, $X$ ranking is the reverse of $Y$.

- $\tau = 0$: Implies that $X$ and $Y$ are independent.

The statistical Kendall's Tau-b coefficient significance given by $p$-value is

$$p = \frac{3\tau\sqrt{n(n-1)}}{\sqrt{2(2n+5)}} \qquad (6)$$

We refer the reader to [22,23] for a comprehensive treatment of the statistical and mathematical properties of Kendall's Tau-b coefficient.

*Classifiers*

Linear discriminant, quadratic discriminant and support vector machine (SVM) are machine learning techniques used in pattern recognition problems in order to make predictions or decisions about the input data. The process of predicting the class of given input data is according to the targets, labels or categories during the classification stage. In our case, these learning techniques use the input vector $[\tau, p] \in \mathbb{R}^2$ in order to predict between two classes namely: spike-and-wave and non-spike-and-wake, both training and classification stage. In these techniques the sensibility and the specificity are estimated by using the classical ROC analysis [24]. See [25] for deeper information about these techniques.

**Results and Discussion**

We evaluated the proposed methodology using 22 EEG monopolar 256 Hz channels of one patient of study during a long-time recording, and 300 spike-and-wave epileptiform discharges from the database presented above in subsection *Database*. Note that, the long-time signals were recorded during a sleep period of 8 hours. All the epochs of the study were selected by the physician. Therefore, each epoch had the beginning and end of each spike-and-wave epileptiform discharge, which we used as ground truth.

For illustration, Figure 3. shows some examples of the annotated medical data (a, c, e) and candidates for our method (b, d, f) in blue colors with respect to the SWD pattern in red color. The annotated medical data was chosen in a medical context, looking at the signal in the time domain, see Figure 1. In (b) example, we can see the similar morphology. In (d) example a wave is detected, but not the spike. It is a false detection, and in (f) example a false detection is found.

In total 3080 $\tau$ and $p$-values were estimated in 140 segments by each channel. The obtained $\tau$ and $p$-values for each segment per channel were corroborated with the medical annotations by visual inspection. Based on this analysis, a threshold= 0.5 and $p$-value= 0.05 were selected in order to perform a ROC analysis [24].

The threshold value was selected because values that were higher than this value were good candidates to be spike-and-waves, considering that the Kendall's Tau-b coefficient is a nonparametric correlation analysis; and the p-value was estimated using the equation (6). Therefore, the Tau-b statistic value was assigned to the annotated data. Next, these were compared with the threshold value; and later added to the confusion matrix in order to assign a specificity and sensitivity metric value.

The percentage of correct classifications was analyzed only in terms of high Specificity, rule in (SpPIn) [26-28], because it is known that, the patient of study has spike-and-wave discharges and in this preliminary study it was important to know if the proposed method detects the SWD pattern. The values obtained were: 14% sensitivity (True positive rate), 83% specificity (True negative rate) evaluated with all 300 spike-and-waves from database in 3080 segments. Specificity values range from 0.81 to 0.84 while sensitivity values range from 0.12 to 0.16 following a 95% confidence interval.

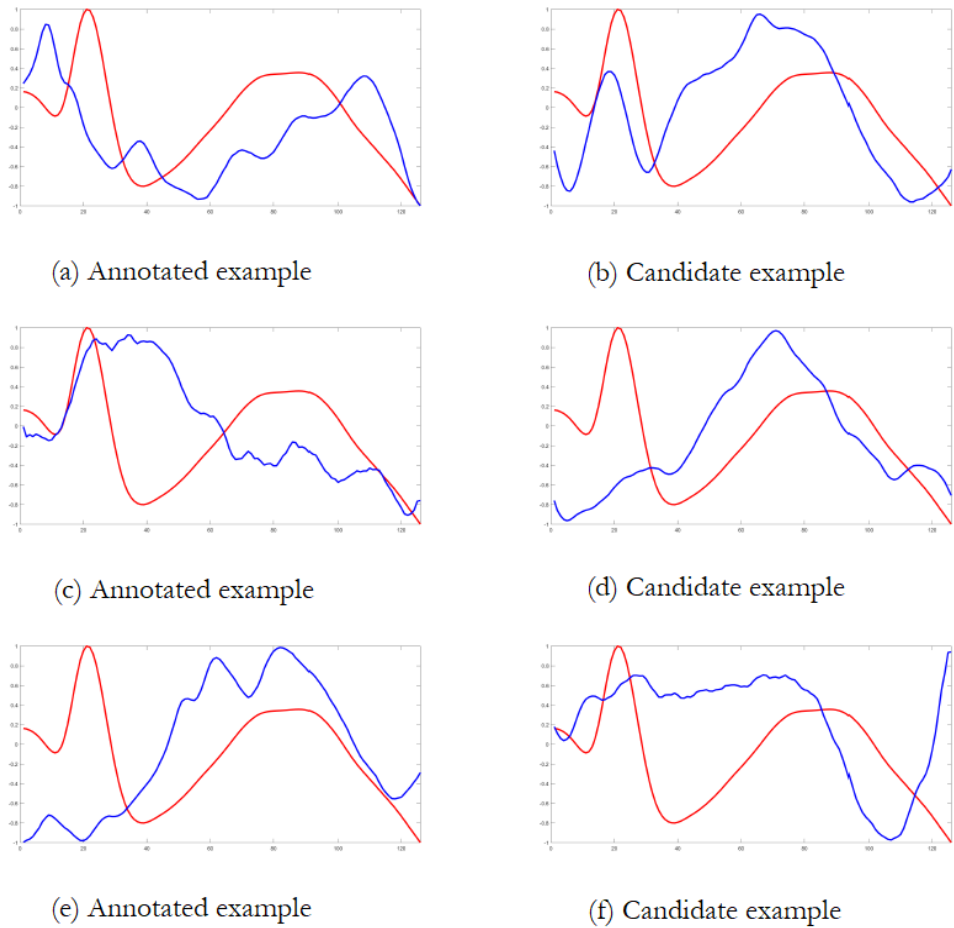

**Figure 3.** Left column are annotated medical data (a), (c) and (e). Right column are candidates for our method (b), (d) and (f). We can see how the EGG waveforms (blue color) have similar morphology with the SWD pattern (red color). In (b) example, we can see the similar morphology. (d) and (f) examples, are false detections. In (d) a wave is detected, but not the spike and in (f) the waveform is different.

If only 10 extracted patterns from each new patient are analyzed and included into the database, SpPIn results improve by 11%; with 94% specificity with values ranges from 0.93 to 0.94 for the 95% confidence interval. This presented us with our next hypothesis: spike-and-wave epileptiform discharge detection must be focused on patient-specific spike-and-wave detection. With this hypothesis, three supervised classifiers were tested with leave-one-out cross-validation technique from vector $[\tau, p] \in \mathbb{R}^2$, obtaining the same SpPIn results (94% specificity) only with the 10 SWD patterns: linear discriminant, quadratic discriminant and lineal Support Vector Machine (SVM). This suggests that with our method is possible train a classifier and detect with a small dataset, similar to [29]. Our results are promising in comparison to other methods of the state-of-the-art that reports their SpPIn[2] by using harmonic analysis with SpPIn of 97%, or mixing the discrete cosine transform with Daubechies wavelets with SpPIn 90.7% [3], or applying the Morlet wavelet energy with SpPIn of 98.7% [30].

The main limitation was defining the sliding time-window and the perfect overlap of epochs because epileptic signals have a high dynamic. The main advantage is that, it is possible to train a classifier and detect with a small dataset. But this remains an open issue.

Future work will focus on a deep evaluation of the proposed approach with a large group of patients: to determine the amplitude difference between signals in parallel brain regions, to estimate the $l_1$-norm coupled with a filtering stage to outliers control [31], to apply an adaptive signed correlation index [14], to improve the sensitivity in new patients with regularization techniques [32] and to tune parameters with information theory [33] or ensembles from multiple models [34].

**Conclusions**

This work presented a new spike-and-wave detection method to detect epileptiform discharge patterns in EEG signals. The method is based on Kendall's Tau-b coefficient which captures the statistical relationship between two waveform signals involved in the bivariate structure in order to quantify and detect the morphological similarity or concordance between signals. The proposed methodology was demonstrated on 3080 EEG segments of 300 monopolar 256 Hz spike-and-wave discharges database from Fundación Lucha contra las Enfermedades Neurológicas Infantiles (FLENI), which suggests that the proposed algorithm is a powerful tool for detecting seizures in epileptic signals in terms of high Specificity, rule in (SpPIn), 94% for patient-specific SWD detection and 83% for a general SWD detection.

**List of abbreviations**

EEG: Electroencephalography.
SpPIn: high Specificity, rule in.
SWD: spike-and-wave discharge.

**Conflict of Interest**

Authors declare that they have no conflict of interest.

**Authors' Contributions**

AQR carried out the aim of research and the design of experiment. AQR, CC, JE and LH carried out the experiments and participated in the design of the study and performed the statistical analysis. VM, CD and AQR designed the database with the medical annotations. All authors coordinated and helped to draft the manuscript. All authors read and approved the final manuscript.

*Antonio QUINTERO-RINCÓN, Catalina CARENZO, Joaquín EMS, Lourdes HIRSCHSON,*
*Valeria MURO, Carlos D´GIANO*